\documentclass[a4paper, 10pt, conference]{ieeeconf}      
\usepackage{FG2023}
\usepackage{times}
\usepackage{latexsym}
\usepackage{multirow}
\usepackage{amsmath}
\usepackage{microtype}
\usepackage{graphicx}
\usepackage{inconsolata}
\usepackage{booktabs}
\usepackage{float}
\DeclareMathOperator*{\concat}{concat}

\FGfinalcopy 

\IEEEoverridecommandlockouts                              
\def\FGPaperID{119} 

\title{\LARGE \bf
Face-to-Face Contrastive Learning for Social Intelligence Question-Answering
}

\author{\parbox{16cm}{\centering
    {\large Alex Wilf$^{1,*}$, Martin Q. Ma$^{1,*}$, Paul Pu Liang$^2$, Amir Zadeh$^{3,1}$, Louis-Philippe Morency$^1$}\\
    {\normalsize
    $^1$ Language Technologies Institute, Carnegie Mellon University\\
    $^2$ Machine Learning Department, Carnegie Mellon University \\
    $^3$ Amazon Alexa AI}}
    \thanks{$^*$ Equal contribution}
}

\begin{document}

\ifFGfinal
\thispagestyle{empty}
\pagestyle{empty}
\else
\author{Anonymous FG2023 submission\\ Paper ID \FGPaperID \\}
\pagestyle{plain}
\fi
\maketitle
\thispagestyle{plain}
\pagestyle{plain}

\begin{abstract}

Creating artificial social intelligence – algorithms that can understand the nuances of multi-person interactions – is an exciting and emerging challenge in processing facial expressions and gestures from multimodal videos. 
Recent multimodal methods have set the state of the art on many tasks, but have difficulty modeling the complex face-to-face conversational dynamics across speaking turns in social interaction, particularly in a self-supervised setup. In this paper, we propose Face-to-Face Contrastive Learning (F2F-CL), a graph neural network designed to model social interactions using \textit{factorization nodes} to contextualize the multimodal face-to-face interaction along the boundaries of the speaking turn. With the F2F-CL model, we propose to perform contrastive learning between the factorization nodes of different speaking turns within the same video. We experimentally evaluate our method on the challenging Social-IQ dataset and show state-of-the-art results.

\end{abstract}
\section{Introduction}
\label{sec: introduction}
Effectively modeling face-to-face social interaction is an important task that presents key challenges for automatically understanding body language \cite{de2019language}, in part because the benchmarks for the task are relatively low-resource \cite{zadeh2019social} . Recent years have shown the effectiveness of self-supervised learning combined with expressive model architectures such as graph neural networks and transformers \cite{hu_heterogeneous_2020} to improve state-of-the-art performance on low-resource tasks \cite{tran2020cross}.  In this paper, we consider one self-supervised approach in particular, contrastive learning, and ask \textit{How can we train models to understand face-to-face social interaction using self-supervised contrastive learning?}
\\

\begin{figure}[ht]
\includegraphics[width=0.95\columnwidth]{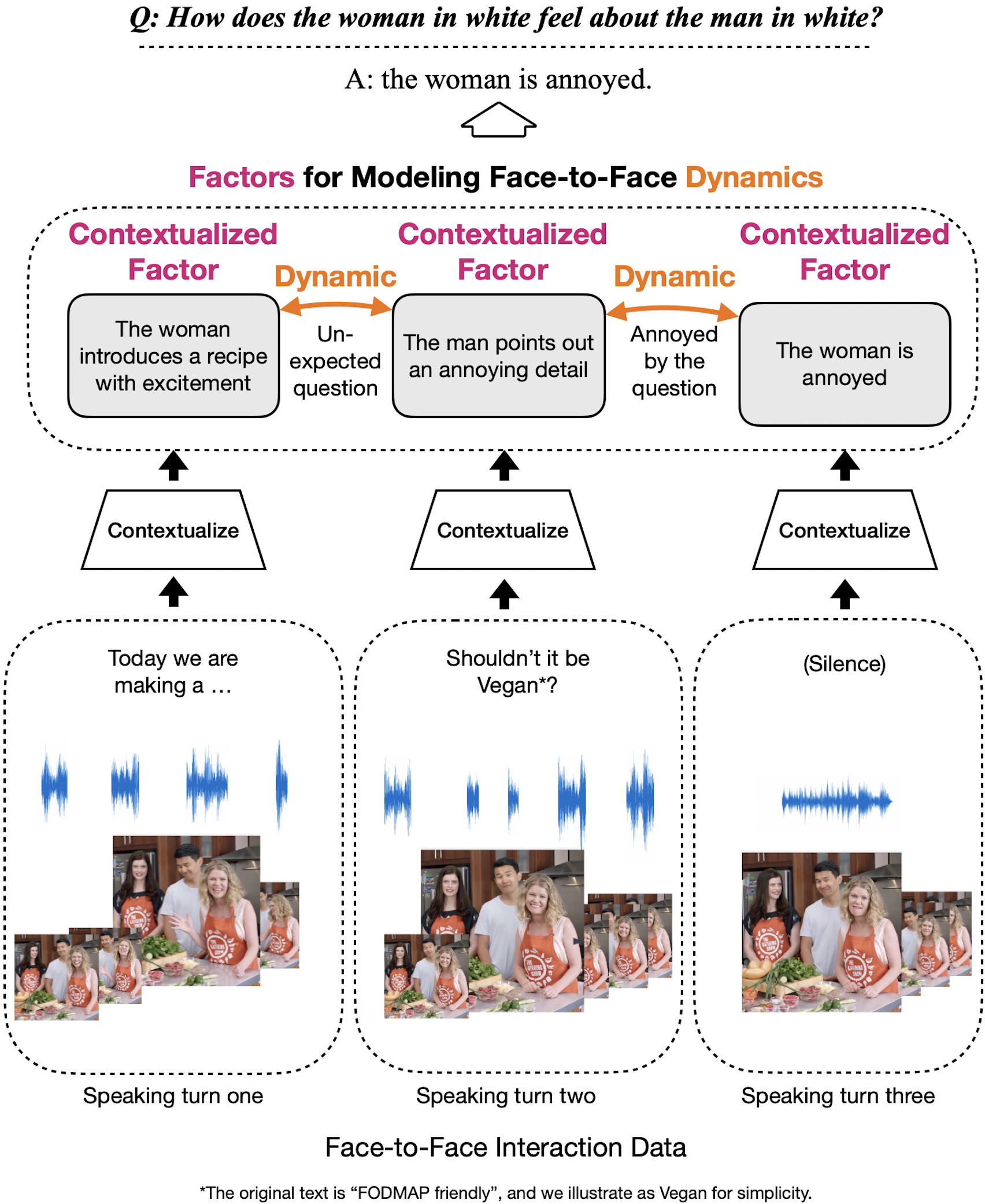}
  \centering
  \caption{An example approach for modeling social interaction. It first contextualizes face-to-face interaction within each speaking turns into some contextualized latent factors, and then models the cross speaking-turn dynamics. This approach nicely captures the face-to-face conversational dynamics by leveraging speaking turns.}
\label{fig:motivation}
\end{figure}
\vspace{-2mm}

Contrastive learning is an approach to self-supervised learning in which the model is encouraged to encode ``positive pairs'' of inputs in a similar way and ``negative pairs'' differently \cite{oord2018representation, chen2020simple, he2020momentum}. An effective contrastive learning setup for this task will require two components: a way of defining positive and negative pairs that provides a strong learning signal and an architecture that can effectively model face-to-face social interaction. To do the latter, a model must both understand the conversational dynamics of each speaking turn and be able to understand interactions across speaking turns. However, despite efforts to improve contrastive learning for videos \cite{zeng2021contrastive, kuang2021video},  there has not been a contrastive framework that effectively models social interaction by using the rich conversational dynamics of videos. A good approach should contextualize information within speaking turns, meanwhile modeling the dynamics across turns. We illustrate the essence of this approach in Figure \ref{fig:motivation}, where each contextualized factor could capture speaking-turn information and interact with other contextualized factors to model conversational dynamics.

In this paper, motivated by the idea of contextualization and interaction, we propose \textbf{Face-to-Face Contrastive Learning}, a graph-based approach to model multimodal social interaction using contrastive learning. Our method first encodes the multimodal videos into fully connected graphs based on speaking turn boundaries, and then contextualizes each speaking turn using a novel node structure, which we call factorization nodes. With this structure, we perform contrastive learning by explicitly modeling the interactions between factorization nodes. We treat the factorization nodes from the same speaking turn as the positive pair, and those from different speaking turns as the negative pairs. The factorization nodes must contextualize information within each speaking turn to perform well on the contrastive learning objective, and the interactions between different factorization nodes can effectively capture face-to-face conversational dynamics across speaking turns. 

Empirically, we evaluate on the publicly available Social-IQ dataset~\cite{zadeh2019social} comparing with both supervised and self-supervised baselines, and achieve state-of-the-art results on the social interaction question-answering task.

\section{Related Work}

\paragraph{Social Intelligence} A benchmark called Social-IQ (Social Intelligence Queries)~\cite{zadeh2019social} provides a diverse set of videos and evaluates models' ``social intelligence'' through many ``why'' and ``how'' questions related to social interaction. We run our experiments on this benchmark because it is more focused on social interaction than other video QA datasets such as TVQA~\cite{lei2018tvqa} and MovieQA~\cite{tapaswi2016movieqa}. Graph neural networks have not been used for this task before, so to thoroughly evaluate our claims, we provide results not only our method, but also for ablations of our method that represent common GNN approaches to other video QA tasks.

\paragraph{Multimodal Graph Network} Graph neural networks (GNN)~\cite{gori2005new, scarselli2008graph} have been proposed to encode graph-structured data, and have been applied on multimodal tasks such as Visual Dialog~\cite{guo2020iterative} and VQA~\cite{hudson2019gqa}. \cite{hu_heterogeneous_2020} show that multimodal transformers can be considered a special case of fully connected GNNs, including large multimodal transformers such as MERLOT~\cite{zellers_merlot_2022} and UNITER~\cite{chen2020uniter}. Most similar to our work is \cite{yang2021mtag}, which uses a graph neural network to model cross-modality, within-modality, and temporal connections, but differs from our strategy in our factorization strategy and contrastive learning setup.

\paragraph{Self-Supervised Contrastive Learning}
Self-supervised learning attempts to create ``labels'' from data (e.g. masking) or the training process itself (e.g. contrasting) to pretrain representations~\cite{jaiswal2021survey} and uses the learned representations for downstream tasks \cite{chen2020big, he2020momentum, lan2019albert, oord2018representation}. A well-studied method of self-supervised learning is contrastive learning~\cite{oord2018representation, he2020momentum, chen2020simple}, which tries to construct pairs of samples that should be represented similarly (termed positive pairs) and pairs of samples that should be represented differently (termed negative pairs) and learn a model that distinguishes the positive pair from the negative pairs. Contrastive learning methods are widely used in graph representation learning. For example, \cite{velivckovic2018deep} treats node-level and graph-level representations of the same graph as positive pairs, and different graphs as negative pairs. \cite{you2020graph} stochastically perturbs the edges and nodes on the same graph to create a positive pair and treats random graphs as negative pairs.

The most similar to our work on contrastive learning for videos are \cite{zeng2021contrastive} and \cite{kuang2021video}. \cite{zeng2021contrastive} learns representations of videos by defining a loss considering positive pairs as different augmentations of the same temporal segment within a single video, and negative pairs as different temporal segments. Our work differs from this approach in that instead of defining segments with a fixed window size, we assign segment boundaries based on speaking turns. \cite{kuang2021video} also considers temporal segments within videos, but instead of contrasting at the temporal segment level as we do, contrasts by collating frames drawn at random from different temporal segments.
There are many other works that perform contrastive learning on videos \cite{wu2021contrastive, yuan2022contextualized, ding2022motion, xu2021rethinking}, but we have found none that contrasts based on dynamic and semantically meaningful temporal boundaries (such as speaking turn).

\paragraph{Factorization in Multimodal Networks}  
\cite{song2013action} groups neighboring frames of sequences that have similar semantic information and uses group-level representation for event summarization. 
\cite{niu2017hierarchical} models the hierarchical relations between sentences and phrases and between whole images and image regions to learn a joint embedding for image captioning. 
Many more networks have used factorization effectively on a variety of multimodal tasks, such as \cite{chen2022scaling,wu2020learning,tsai2018learning,zadeh2019factorized}.

\section{Face-to-Face Contrastive Learning}
In this section, we describe the proposed Face-to-Face Contrastive Learning (F2F-CL) to model social interaction. Our method consists of two main steps: (1) encoding multimodal video into graphs and encoding speaking-turn level information into factorization nodes and (2) performing contrastive learning using the factorization nodes.

\subsection{Problem Definition}
We define the task of modeling social interactions given the multimodal video, including textual input $T$, visual input $V$ and acoustic input $A$, and questions and answers regarding the social interactions in the video.
The task is to model inter-speaker interaction to correctly answer the questions.  Following prior work \cite{yang2020mtag, zadeh2019social}, we feed each modality into a modality-specific encoder to produce unimodal features with the same hidden dimensions (details in Section~\ref{subsec: impl}). Then, we partition the processed features into groups based on speaking turn boundaries \cite{kuchaiev2019nemo}. We leave an investigation of the relationship between speaking turn boundary quality and our algorithm's performance to future work.

\begin{figure}[t]
\includegraphics[width=\columnwidth]{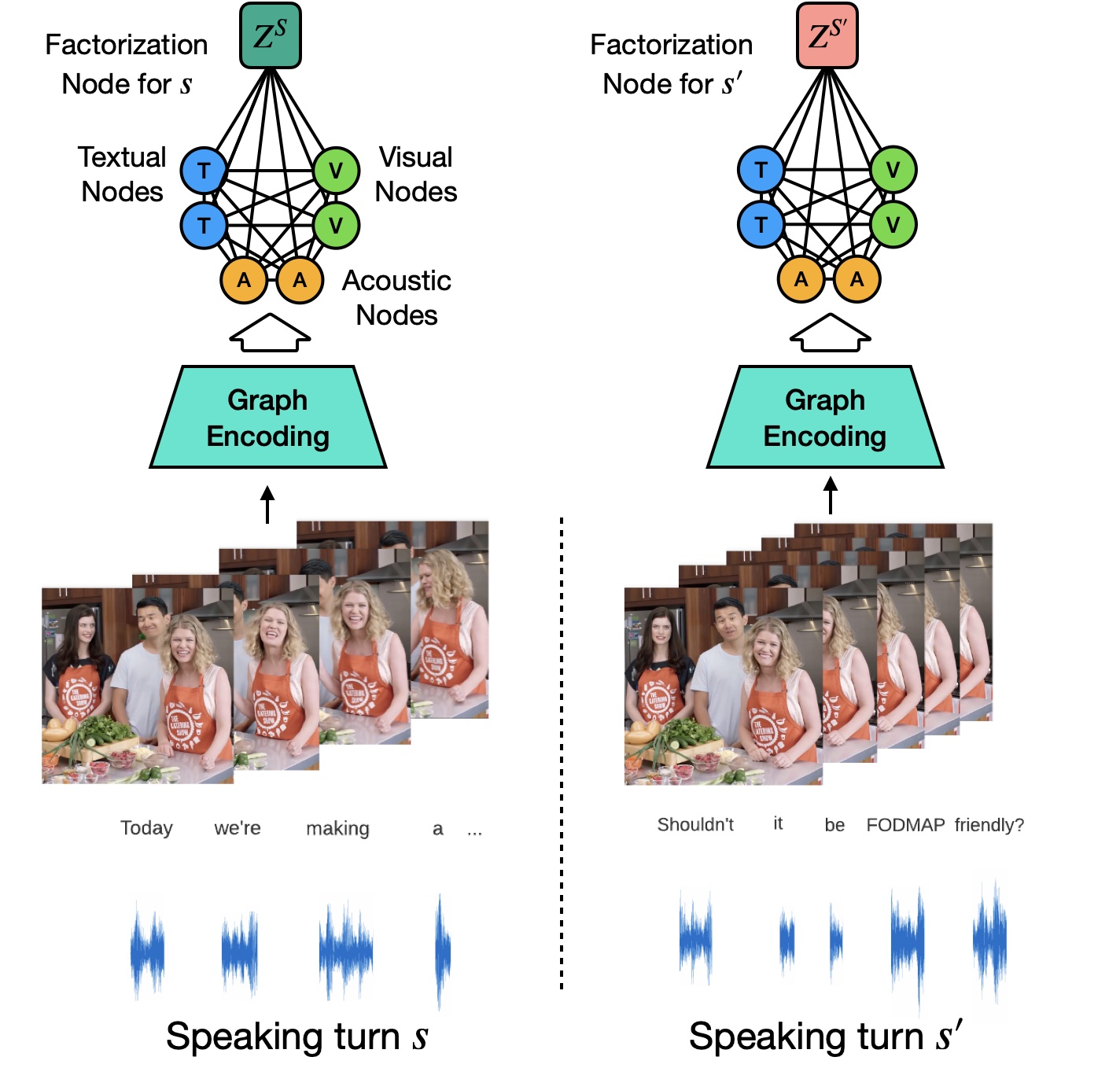}
  \centering
  \caption{Constructing speaking-turn graphs. We first segment the videos by speaking turn, and construct a fully connected graph for every speaking turn where a node is a modality at one time step. Then the factorization node extracts information from all modalities in the speaking turn. Each factorization node represents one speaking turn.}
\label{fig:speaking_turn}
\end{figure}

\subsection{Graph Encoding and Factorization}
\label{sec:graph_construction}

We create speaking-turn graphs based on unimodal features. Formally, we first create nodes denote the visual ($v$), textual ($t$) and acoustic ($a$) nodes of the speaking turn $s$ as $\{v_i^s\}_{i=1}^{N_{v}}$, $\{t_i^s\}_{i=1}^{N_{t}}$ and $\{a_i^s\}_{i=1}^{N_{a}}$, respectively, where each node is a vector representing one modality at one temporal step. The numbers of modality nodes $N_v, N_t, N_a$ are possibly different for each modality. For each pair of nodes, there is a connecting edge representing their interaction. We denote the edge connecting $t_i^s$ to $v_j^s$ by $e^s_{t_i, v_j}$. Each edge is directed ($e^s_{t_i, v_j}$ is different from $e^s_{v_j, t_i}$). Since the graph is fully connected within each turn, there will be $N_{e} = {{N_{v} + N_{t} + N_{a}} \choose 2}$ edges for each turn. For each edge, there will be a normalized attention score $\alpha$ and an unnormalized attention score $\beta$, whose details we will introduce soon.

Next, we connect each node to a single ``factorization node'' $z^s$, which pools representation information from the whole speaking-turn graph by fully connecting to all other nodes in the graph. The edge connecting $z^s$ and $v_j^s$, for example, will be $e^s_{z, v_j}$. We illustrate the overall architecture in Figure \ref{fig:speaking_turn}.

Then, we perform attention updates using the node embeddings and edge weights in the graphs. There are three steps: the first is to compute the raw attention score $\beta^{[h]}$ between two nodes (we use $t_i$ and $v_j$ as an example, neglecting the $s$ superscript for simplicity because all operations are within the same speaking turn $s$): 

\begin{equation}
        \beta^{[h]}_{t_i, v_j} = \text{LeakyRelu}(e^s_{t_i, v_j} \cdot [t_i \Vert v_j]),
\end{equation}
where $[\cdot||\cdot]$ denotes the concatenation of node representations. The $[h]$ index is the head index of multi-head attention. The next step is then to calculate the normalized $\alpha^{[h]}$ of $\beta^{[h]}_{t_i, v_j}$ on all neighbors of $t_i$ using softmax:
\begin{equation}
\alpha^{[h]}_{t_i, v_j}
=\frac{\exp ( \beta^{[h]}_{t_i, v_j} )}{\sum_{k\in {N}_{t_i}} \exp ( \beta^{[h]}_{t_i,k} )},
\end{equation}
where $N_{t_i}$ denotes all the neighboring nodes of $t_i$. In our fully connected case, $N_{t_i}$ includes all nodes except $t_i$. In the last step, node $t_i$ is updated according to the attention scores:
\begin{equation}
t_i = \concat\limits_{h=1}^H(\sum_{k\in {N}_{t_i}} \alpha^{[h]}_{t_i,k} t_i)
\end{equation}
where $H$ is the number of total attention heads. After this step, $t_i$ encodes heterogeneous modal-temporal interactions between $t_i$ and its neighbors. We perform such updates for all other modality nodes, $\{v_i^s\}_{i=1}^{N_{v}}$, $\{t_i^s\}_{i=1}^{N_{t}}$ and $\{a_i^s\}_{i=1}^{N_{a}}$. More importantly, the same also applies to the factorization node $z$, where its update will be based on the neighboring nodes ${N}_{z}$ and the attention scores between $z$ and its neighbors:

\begin{equation}
z = \concat\limits_{h=1}^H(\sum_{k\in {N}_{z}} \alpha^{[h]}_{z,k} z)
\end{equation}

Because ${N}_{z}$ contains all the modality nodes in speaking turn $s$, $z$ contextualizes all the modality information in speaking turn $s$. Such updates are performed on all speaking turns so that $z^s$ contextualizes the speaking turn $s$.

\subsection{Contrastive Learning with Factorization Nodes}
\label{sec:ssl_method}

\begin{figure}[ht]
\includegraphics[width=\columnwidth]{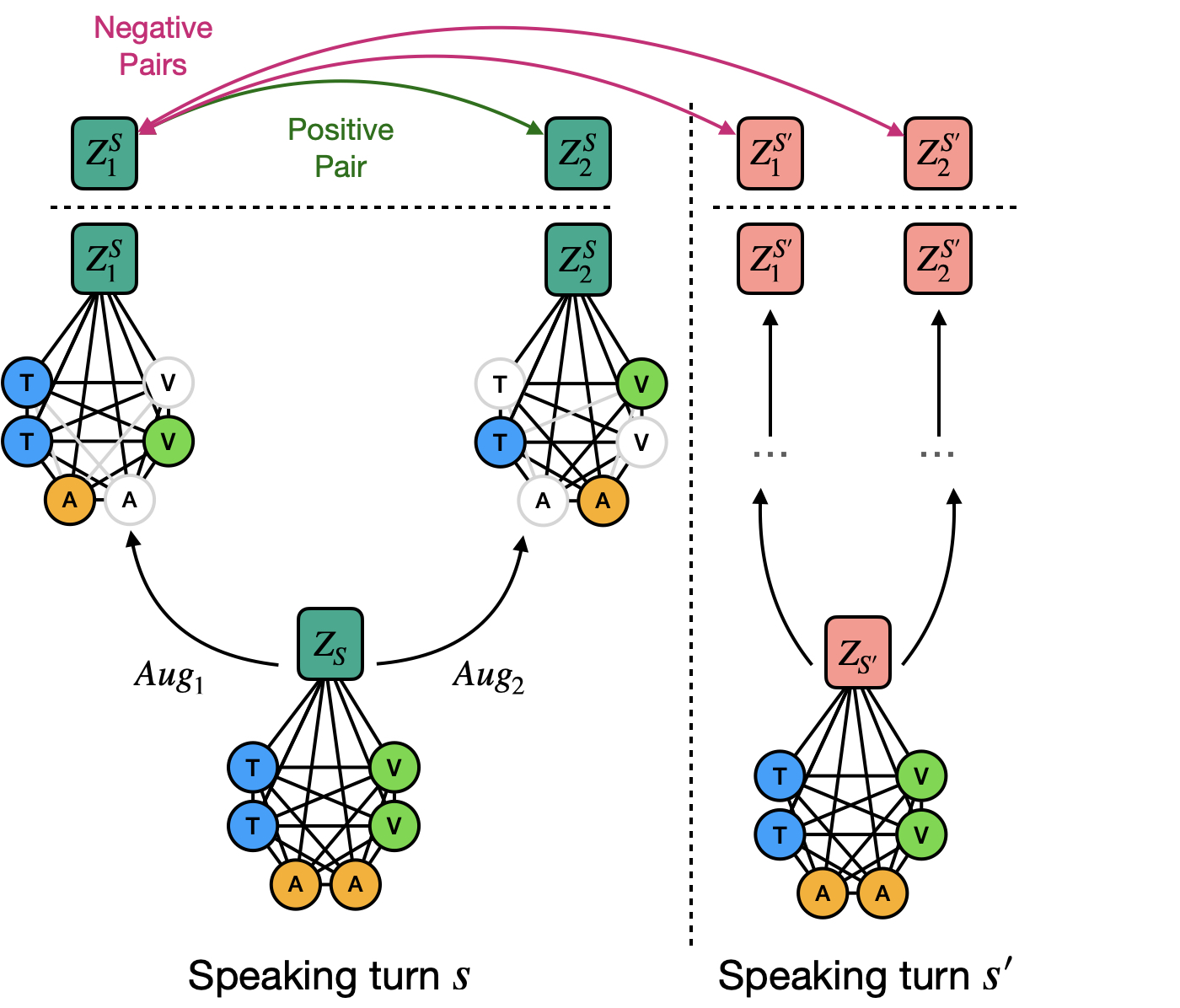}
  \centering
  \caption{Contrastive learning example. $Aug_1$ and $Aug_2$ are two random augmentations. One speaking-turn graph is augmented to create two augmented graphs, and contrastive learning is performed on the factorization nodes. Nodes from the same speaking turn are a positive pair, and nodes from different speaking turns are negative pairs. }
\label{fig:contrastive}
\end{figure}

Next, we introduce the self-supervised contrastive learning step. Contrastive learning \cite{chen2020simple, he2020momentum} creates positive and negative pairs from data and uses a contrastive objective to pull together embeddings of positive pairs and push away embeddings of negative pairs. The first step is to create positive and negative pairs. We demonstrate the overall approach in Figure \ref{fig:contrastive}.

\paragraph{Creating positive and negative pairs} Following \cite{chen2020simple, you2020graph}, we define \textit{stochastic data augmentations} on graphs as creating realistic new graphs by applying a certain transformation without affecting the semantics label of the graph. We want to augment each speaking turn graph to create two different but related graphs. Given a speaking turn graph $\mathcal{G}$, we discuss the following four types of graph augmentation technique to create \textit{two} graphs for each speaking turn:
\begin{enumerate}
    \item \textbf{Node Dropping} drops a ratio of the nodes in the graph $\mathcal{G}$, together with their connecting edges with other nodes in the graph. It assumes that the missing part of the nodes in graph $\mathcal{G}$ does not affect its semantic meaning.
    \item \textbf{Edge Perturbation} randomly changes the edges in $\mathcal{G}$ by adding or dropping edges by a certain ratio. It assumes that the semantics of $\mathcal{G}$ is reasonably robust to edge connection variations. 
    \item \textbf{Node Masking} sets the embeddings of some nodes to zero vectors. The underlying assumption is that missing nodes do not affect the overall graph information much. 
    \item \textbf{Subgraph Sampling} samples a subgraph from $\mathcal{G}$ using a random walk \cite{you2020graph}. It assumes that the semantics of $\mathcal{G}$ can be much preserved in its local (partial) structure.  
\end{enumerate}
Node dropping, edge perturbation, and node masking follow uniform distributions to select nodes or edges. Note that we do not perform augmentation on the factorization nodes or their corresponding edges, as they need to extract information from all other nodes. 

Next, we perform contrastive learning. After graph augmentation, every speaking turn now has two augmented graphs, each with a unique factorization node. We define positive pairs as two randomly augmented graphs from one speaking turn graph, while negative pairs are graphs from different speaking turns. We use a contrastive objective which takes into account only the factorization nodes, not the whole graph. Assuming that we have $N_S$ speaking turns, we denote factorization nodes from a positive pair as $z^s_{1}$, $z^s_{2}$. Inspired by the popular InfoNCE loss \cite{oord2018representation, chen2020simple}, we formulate our loss below for the $s$-th speaking turn: 
\begin{equation}
\mathcal{L}_{ssl}= - \log \frac{e^{sim(z^s_{1},z^s_{2}) / \tau}}{\sum_{s' \neq s}^{N_{S}} \sum_{i=1}^{2} e^{sim(z^s_{1},z^{s'}_{i}) / \tau}},
\end{equation}
where $sim(\cdot)$ is a cosine similarity function and $\tau$ is a temperature hyperparameter. The final loss is computed for all positive pairs of factorization nodes at different speaking turns. 

\paragraph{Why does factorization help?} The key benefit of factorization is improving contrastive learning to model social interactions. By factorizing based on speaking turns, we can augment each speaking turn to create multiple positive pairs and negative pairs within a single video. Each video could generate many positive and negative pairs, compared to the case without factorization where only multiple videos can generate positive and negative pairs. This scales the number of positive and negative pairs by the average number of speaking turns per video, increasing the data efficiency of our method.

\subsection{Supervised Fine-tuning on Question-Answering}
To evaluate the representation quality of the model after self-supervised pretraining we perform a supervised fine-tuning, where we freeze the learned model and fine-tune a two-layer network that considers the hidden representation of the graph neural network. In our case, because we evaluate on Social-IQ \cite{zadeh2019social}, which is a data set of questions and answers, we use the questions and answers as labels to perform the fine-tuning. 

To incorporate the QA representations, the question and answer sequences (both correct and incorrect) are first encoded by separate LSTM networks as in~\cite{zadeh2019social}, and the representations of questions, correct answers, incorrect answers, and multimodal input are all contextualized in a graph convolution operation. We use the updated Graph Attention Layer (GATv2) \cite{brody_how_2022} as our graph convolution to operate on all edges, including edges connecting two modality nodes, edges connecting question and answer (correct and incorrect) nodes, and edges connecting modality and Q/A nodes.  We run a fixed number of convolutional layers over the graph to create a contextualized representation vector for each pair of question and answer. Our fine-tune layer predicts, using contextualized representations for each question and answer pair, a number between zero and one (zero means the answer to the question is wrong, while one means the answer to the question is correct) by a sigmoid activation. We randomize which position the correct answer is in and train with a mean squared error loss between the predicted and correct answer indexes.

\section{Experimental Methodology}
\label{sec: exp}

Our experiments evaluate the quality of the representations learned through our novel contrastive approach. In the following, we describe the dataset as well as the baselines.

\subsection{Experimental details}
\label{subsec: experiment_details}

\paragraph{Dataset} We are using the Social-IQ dataset \cite{zadeh2019social}.  Social-IQ proposes using a diverse set of videos to reason about human social interaction. Videos often contain a social scene, where different people interact with each other. The dataset evaluates models' performance on certain types of ``why'' and ``how'' questions about the interaction between people in the video based on the social scene, which is well below human performance ($68.8\%$ vs. $95.08\%$)~\cite{kumar2020mcqa}, even though the models perform well in other multimodal QA datasets. We split the dataset along the standard splits defined in the Social-IQ dataset SDK and evaluate our algorithm on the held out validation set.

\paragraph{Data Preprocessing} We use as input modalities BERT \cite{devlin2018bert} representations for text, COVAREP representations for audio, and features from a large pretrained image transformer called BEIT \cite{bao2021beit}. We use the ``aligned'' version of the Social-IQ dataset, meaning that we average representations of higher frequency modalities across timestamps denoting the start and end of each word boundary, as in\cite{zadeh2017tensor, zadeh2019social}. The number of modality nodes for each graph is equivalent across modalities, where, for example, the first vision node represents the average of the BEIT vectors for the vision frames corresponding to the timestamps for the first word.

\paragraph{Implementation Details}
\label{subsec: impl}
Our hyperparameter choices are that we use the AdamW optimizer \cite{devlin2018bert} with learning rate $.001$, shared dimensionality across all graph nodes of $80$, and 2 graph attention layers, each containing graph convolutions for each modal-temporal pair. We train with a batch size of 15, sequence length 25, and 25 max epochs (although the network often converges well before then), reporting the max validation accuracy.

For supervised training (baselines we will introduce next), we model the modality-modality components of each graph similar to the fashion in \cite{yang2021mtag}, and  implemented in the Pytorch-Geometric package~\cite{fey2019fast}.  We add the following edges and associated graph convolutions to the graph: the edge connecting questions, the edge connecting answers, the edge connecting questions and answers, and the edge connecting factorization nodes $z^s$ to the question and answer nodes.  Our training takes ~3 hours on a single NVIDIA 3060Ti GPU. The code will be released soon.

\subsection{Baselines}

In the following, we introduce two types of baselines: self-supervised baselines and supervised baselines. The self-supervised baselines are contrastive baselines without factorization nodes. The supervised baselines are the prior state-of-the-art baselines on Social-IQ.

\subsubsection{Self-Supervised Baselines}
The self-supervised baselines we test perform the same graph augmentations and contrastive pre-training but with two different ways of creating factorization nodes: none (one video-level factorization node created by averaging all node representations as in \cite{yang2021mtag}), and mean (segment the graph, create factorization nodes for each segment using mean readout of nodes in the subgraph).

\textbf{Video-Level Contrastive, No factorization nodes.} This approach creates positive and negative pairs from multiple videos. Each positive pair has two graphs created based on the whole \textit{video}, not on a single speaking turn. Negative pairs are graphs created from different videos. In other words, this approach uses a fully graph to represent each video, augments the graph to create two slightly different graphs for each video, and performs contrastive learning on augmented video-level graphs.

\textbf{Speaking-Turn-Level Contrastive, Mean of nodes.} This approach is an ablation of our method. This baseline first segments the graph based on speaking turns, similar to ours, and then directly averaging all node embeddings of the speaking turn graphs as the representations of the graph. The representation is similar to our proposed factorization node in the sense that both contextualize the whole graph. The difference is that the baseline does not parametrize the connections between the graph representation and the modality nodes, whereas the proposed factorization nodes do.


\subsubsection{Supervised Baselines}
Below we also introduce several other baselines that achieved prior state-of-the-art results on Social-IQ. All of them are supervised methods. We consider three published baselines for the Social-IQ task as well as two versions of our model. The baseline models that we compare are listed below.

\textbf{TMFN}~\cite{zadeh2019social}, which uses Tensor Fusion \cite{zadeh2017tensor} for multimodal fusion in the recurrent stages of the MFN network \cite{zadeh2018memory}.

\textbf{MCQA}~\cite{kumar2020mcqa}, which proposes to perform cross-modal alignment and multimodal context-query alignment to solve the social intelligence task.

\textbf{MTAG}~\cite{yang2021mtag}, which uses a fully connected modal-temporal graph neural network to contextualize nodes, then uses a mean readout operation to arrive at a single latent scene vector which is used for classification. MTAG was originally applied to classification tasks; to apply it to QA, we reimplement it by fusing the graph readout representation with the q/a/i nodes and use a two-layer MLP as in our model.

The supervised baselines, which are versions of our model trained from scratch in a supervised way, are as follows.

\textbf{Video-Level Supervised}: This is an ablation where the question and answer nodes are fully connected to the multimodal graphs at the video level.

\textbf{F2F-CL Supervised}: The F2F-CL model we proposed, but trained directly with question and answer labels. The difference between F2F-CL Supervised and the Video-Level Supervised baseline is that F2F-CL Supervised connects speaking-turn level graphs to the question and answer nodes, while Video-Level Supervised connects Video-Level graphs to the questions and answer nodes.

\section{Results}
\label{sec:results}


\begin{table}[h!]
    \caption{Results on Social-IQ dataset. The proposed F2F-CL outperforms all other self-supervised and supervised baselines and achieves state-of-the-art result.}
    \begin{center}{
    \resizebox{0.8\linewidth}{!}{
    \begin{tabular}{cc}
        \toprule
        Model& Accuracy ($\%$)\\
        \midrule

        \multicolumn{2}{c}{\bf Social IQ QA (Self-Supervised)}               \\ 
        \midrule
        Video-Level Contrastive & 67.90 \\
        Speaking-Turn-Level Contrastive & 70.76 \\
        \textbf{F2F-CL} & \textbf{72.03}\\

        \midrule
        \multicolumn{2}{c}{\bf Social IQ QA (Supervised)}               \\ 
        \midrule
        MFN~\cite{zadeh2018memory} & 64.82\\
        MCQA~\cite{kumar2020mcqa} & 68.80\\
        MTAG~\cite{yang2021mtag} & 65.58\\
        Video-Level Supervised & 70.66 \\
        \bf F2F-CL Supervised & \textbf{72.01} \\
        \bottomrule
    \end{tabular}
    }
    
    \label{tab:main_res}
    }

    \end{center}
\end{table}

First, we find that segmenting the videos into speaking turns helps improve performance.  This is shown in Table \ref{tab:main_res}, as the overall accuracy of contrastive learning at the level of speaking turn (segment level contrastive and F2F-CL) is greater than contrastive learning at video level, even when factorization nodes are defined without parametrization as the mean of nodes in their subgraph. Then we find that the proposed F2F-CL, which relies on factorization on speaking turns with parametrized factorization nodes, achieves the state-of-the-art results on both baselines. 

This result also holds in the supervised case, and surprisingly, we find that the self-supervised setup performs on par with the supervised setup. This is an interesting result, suggesting that the contrastive pretraining within each video with the proposed factorization node structure is a particularly effective method for learning useful multimodal graph representations in a self-supervised way from relatively small amounts of data (compared to many pretraining regimes).

\section{Quantitative Analysis}

In this section, we consider a number of questions to support or challenge our intuition for why F2F-CL learns strong representations from self-supervision.

\paragraph{Factorization Nodes Contain Speaker Information}
We expect that our speaking-turn factorization nodes will learn high level information about the speaking turns they represent. We constructed a dataset to ask the question: \textit{Can we discern, better than random, which speaker a given factorization node refers to?} To do this, we constructed a balanced dataset as follows: for each video in Social-IQ that contains at least two factorization nodes of speaking turn of one speaker ($z^{a}_1$, $z^{a}_2$) and one speaking turn of another ($z^{b}_1$), we add two data points to the dataset: ($z^{a}_1$, $z^{a}_2$; $1$) and ($z^{a}_1$, $z^{b}_1$; $0$). We attempt to learn a simple linear classifier that takes the concatenated input vectors and attempts to distinguish whether they are from the same speaker or not. We train an MLP with two linear layers, one with dimensions $[160,80]$, the other with dimensions $[80,1]$, with ReLU and Dropout between the layers and a Sigmoid layer afterwards for max 30 epochs with early stopping and MSE loss, then we report the validation accuracy. Training and validation sets are the factorization nodes saved from the trained F2F-CL model on the Social-IQ training and validation sets. We find that this model achieved 60.29\% accuracy, above the 50\% accuracy that we would expect randomly.

\paragraph{The Fully Connected GNN Pays Less Attention Across Speaking Turns}
As we describe in Section~\ref{sec: introduction}, we expect that the fully connected MTAG \cite{yang2021mtag} will pay less attention across speaking turns, as it will be trying to learn the ``general idea'' of each speaking turn before reasoning across them. We found support for this idea – attentions for nodes across speaking turns on the validation set were, on average, 36.17\% higher than attentions between nodes within the same speaking turn.

\begin{figure}[H]
\includegraphics[width=0.6\columnwidth]{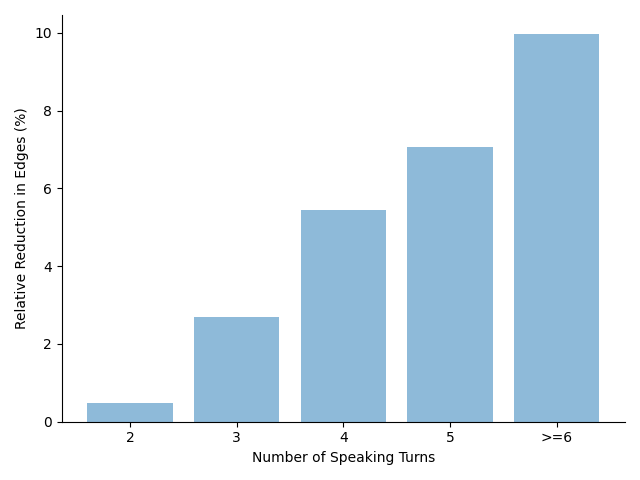}
  \centering
  \caption{Factorization reduces the number of edges in the graph from fully connected GNN.}
\label{fig:num_edges}
\end{figure}

\paragraph{Factorization Reduces the Number of Edges}
We expect that factorization will substantially reduce the number of edges in the graph, which would support our intuition that contextualizing graphs across speaker turns is an effective inductive bias such that cross-speaking-turn interactions can be efficiently modeled by the interactions of factorization nodes. 
We find that factorization reduces the number of edges in multi-speaking-turn graphs. We considered all the videos in Social-IQ for this analysis, as the number of edges is relevant both for the train and validation sets.  Our results are visualized in Figure~\ref{fig:num_edges}.  For videos with 4, 5, and $>=$6 speaking turns, we found a reduction in the number of edges of 5.45\%, 7.08\%, and 9.96\%.

\paragraph{Moderate levels of graph augmentations help the most}
Lastly, we investigate how graph augmentation affects self-supervised learning results. We vary the ratio of each of the four graph augmentations, Node Dropping, Edge Perturbation, Node Masking, and, Subgraph Sampling, while fixing the ratios of the other three augmentations. The default augmentation ratio is set to 0.5. From Table \ref{tab:ssl_ablation}, we observe that F2F-CL benefits from a smaller ratio of augmentations (0.25) on the nodes (Node Dropping and Node Masking) and a moderate ratio of augmentations (0.5) on the edges (Edge Perturbation and Subgraph Sampling) when creating the positive and negative pairs. We attribute this to the fact that the node representation embeddings are much richer, so a medium-to-large (0.5 and 0.75) ratio to drop or mask nodes will likely hurt performances. On the other hand, the edges contains attention weights, so randomly dropping edges will effectively enforcing the model to learn a sparser connections between the factorization nodes and the modality nodes.

\begin{table}[h]
\caption{An ablation analysis of our model reveals that F2F-CL benefits from a smaller ratio of augmentations (0.25) on nodes and a moderate ratio of augmentation (0.5) on edges to create positive and negative pairs in contrastive learning. The numbers below are accuracy ($\%$).  Our full method (which combines augmentations) achieves $72.03\%$ accuracy.}

\centering
    \begin{tabular}{cccc}
        \toprule
        Augmentation & \multicolumn{3}{c}{Ratio} \\
        \cmidrule{2-4}
        & .25 & .5 & .75 \\
        \midrule
        Node Dropping & \textbf{66.81} & 66.79 & 66.76 \\
        Edge Perturbation & 66.14 & \textbf{66.22} & 66.18\\
        Node Masking & \textbf{68.33} & 68.12 & 68.09\\
        Subgraph Sampling & \textbf{68.01} & 67.98 & 67.92\\
        \bottomrule
    \end{tabular}
    \label{tab:ssl_ablation}
\end{table}


\section{Conclusion}
In this paper, we proposed a novel method, Face-to-Face Contrastive Learning (F2F CL), to perform contrastive learning in face-to-face interactions. We leverage the natural conversational dynamics from speaking turns by contextualizing information within each speaking turn into factorization nodes and using the factorization nodes to define positive and negative pairs for contrastive learning. Our method effectively improves the performance of social interaction question-answering and achieves state-of-the-art results on the Social-IQ benchmark.

\section{Acknowledgements}
This material is based upon work partially supported by BMW, National Science Foundation awards 1722822 and 1750439, and National Institutes of Health awards R01MH125740, R01MH096951 and U01MH116925. Any opinions, findings, conclusions, or recommendations expressed in this material are those of the author(s) and do not necessarily reflect the views of the sponsors, and no official endorsement should be inferred.


{\small
\bibliographystyle{ieee}
\bibliography{FG2023}
}

\end{document}